\documentclass[runningheads]{llncs}

\usepackage{graphicx}

\usepackage{amsmath}

\usepackage{cite}
\usepackage{amssymb,amsfonts}
\usepackage{algorithmic}
\usepackage{textcomp}
\usepackage{xcolor}

    \usepackage{float}

\usepackage{msc}

\usepackage[utf8]{inputenc}
\usepackage{amssymb}
\usepackage{dsfont}

\usepackage{algorithm}

\usepackage{bm}

\begin{document}

\title{Learning Optimal Decision Trees from Large Datasets}

\author{Florent Avellaneda}

\institute{Computer Research Institute of Montreal,\\
405 Ogilvy Avenue, Suite 101\\
Montreal (Quebec), H3N 1M3, Canada\\
\email{Florent.Avellaneda@crim.ca}}

\maketitle

\begin{abstract}

Inferring a decision tree from a given dataset is one of the classic problems in machine learning.
This problem consists of buildings, from a labelled dataset, a tree such that each node corresponds to a class and a path between the tree root and a leaf corresponds to a conjunction of features to be satisfied in this class.
Following the principle of parsimony, we want to infer a minimal tree consistent with the dataset.
Unfortunately, inferring an optimal decision tree is known to be NP-complete for several definitions of optimality.
Hence, the majority of existing approaches relies on heuristics, and as for the few exact inference approaches, they do not work on large data sets.
In this paper, we propose a novel approach for inferring a decision tree of a minimum depth based on the incremental generation of Boolean formula.
The experimental results indicate that it scales sufficiently well and the time it takes to run grows slowly with the size of dataset.

\end{abstract}

\section{Introduction}

In machine learning, the problem of classification consists of inferring a model from observations (also called \emph{training examples}) that make it possible to identify which class of a set of classes a new observation belongs to.
When the training examples used to infer a model are assigned to the classes to which they belong, it is called supervised learning.
Many machine learning models exist to solve this problem, such as support vector machines \cite{vapnik1995nature}, artificial neural network \cite{haykin1994neural}, decision trees \cite{quinlan1986induction}, etc.
Because inferring such models is complicated and the number of training examples is generally very large, most existing inference algorithms are heuristic in the sense that the algorithms infer models without any guarantee of optimality.

Although these heuristic-based techniques generally work well, there are always cases where a new example is not correctly recognized by the model.
In the context of critical systems in which errors are not allowed, such models are difficult to use.
This requirement often goes together with the demand that models must also be understandable.
Known as eXplainable AI (XAI), this area consists of inferring models capable of explaining their own behaviour.
XAI has been the subject of several studies in recent years \cite{goebel2018explainable, li2018deep, van2004explainable, angelino2017learning}, as well as several events \cite{ACMFAT, IJCAIXAI}.
One approach to obtain an explainable model is to use decision trees because the reasons for classification are clearly defined \cite{quinlan1986induction}.

In order to obtain accurate and explainable models, we are interested in the inference of optimal decision trees.
The optimality of a decision tree is generally defined by the simplicity of the tree based on the principle of parsimony.
Our chosen simplicity criteria are the depth of the tree and the number of nodes.
In particular, for a fixed maximum depth of the tree, we want to infer a decision tree with a minimum number of nodes that is consistent with the training examples.

Although decision tree inference is a well-studied classic problem, the majority of known algorithms are heuristic and try to minimize the number of nodes without guaranteeing any optimality \cite{quinlan1986induction, salzberg1994c4, kass1980exploratory, breiman1984classification}.
It is because the problem is known to be NP-complete for several definitions of optimality \cite{laurent1976constructing, hancock1996lower}.
In addition, the first algorithms to infer optimal decision trees were ineffective in practice \cite{rokach2008data}.
However, in recent years, several studies have focused on improving the performance of optimal decision tree inference algorithms.

The first series of studies was carried out on a similar problem to ours: inferring optimal decision tree with a given depth such that the total classification error on the training examples is minimized \cite{bertsimas2007classification, bertsimas2017optimal, verwer2019learning}.
For this problem, Verwer and Zhang \cite{verwer2019learning} propose a binary linear programming formulation that infers optimal decision trees of depths four in less than ten minutes.

The studies closest to ours are those of Bessier et al. \cite{bessiere2009minimising} and Narodytska et al. \cite{narodytska2018learning}.
These authors were interested in a particular case of our problem: inferring decision trees with a minimal number of nodes without trying to minimize the depth.
Bessier et al. propose a SAT formulation, but experiments show that the method only works for small models, i.e., trees of about fifteen nodes.
The authors also propose a method based on constraint programming to minimize the number of nodes, but without necessarily reaching the optimal.
Narodytska et al. propose a new SAT formulation that greatly improves the practical performance of optimal decision tree inference.
Thus, with their new formulation, Narodytska et al. were able to build, for the ``Mouse" dataset, a decision tree with a minimum number of nodes in 13 seconds, while this required 577 seconds with the SAT formulation of Bessier et al.
The authors claim that to the best of their knowledge, their paper is the first presentation of an optimal decision tree inference method based on well-known datasets.

In this paper, we propose an even more efficient method than the last one.
Our benchmarks show that we can process the ``Mouse" dataset in only 75 milliseconds.
Moreover, well-known datasets that were considered too large to infer optimal decision trees from them can now be processed by our algorithm.

The paper is organized as follows.
In the next section, we provide definitions related to decision trees needed to formalize the approach.
Section 3 provides a new Boolean formulation for passive inference of a decision tree from a set of training examples.
We propose in Section 4 an incremental way of generating the Boolean formulas which ensures that the proposed approach scales to large datasets.
Section 5 reports several experiments comparing our approach to others.
Finally, we conclude in Section 6.

\section{Definitions}\label{section:def}

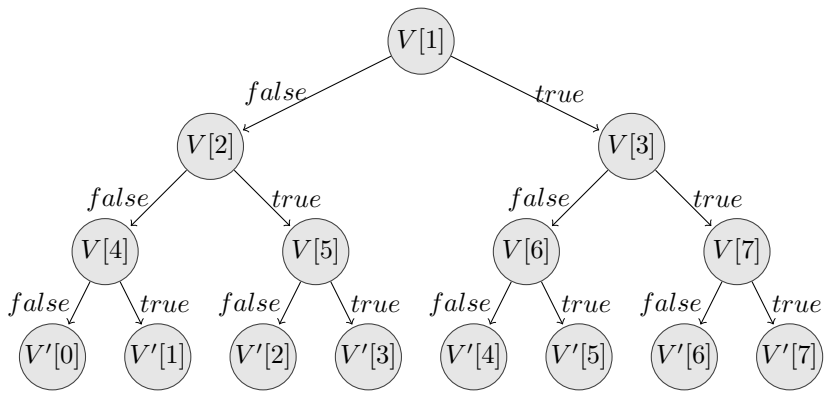
\begin{figure}[h!]
\center
    \begin{tikzpicture}[scale=1.4,shorten >=1pt,->]
    \tikzstyle{vertex}=[circle,fill=black!10,draw=black!75,minimum size=25pt,inner sep=0pt]

    \node[vertex] (Q1) at (0,0) 	{$V[1]$}  ;
    \node[vertex] (Q2) at (-2,-1) 	{$V[2]$}  ;
    \node[vertex] (Q3) at (2,-1) 	{$V[3]$}  ;
    
    \node[vertex] (Q4) at (-3,-2) {$V[4]$}  ;
    \node[vertex] (Q5) at (-1,-2)	{$V[5]$}  ;
    \node[vertex] (Q6) at (1,-2) 	{$V[6]$}  ;
    \node[vertex] (Q7) at (3,-2) 	{$V[7]$}  ;

    \node[vertex] (Q8) at (-3.5,-3) {$V'[0]$}  ;
    \node[vertex] (Q9) at (-2.5,-3)	{$V'[1]$}  ;
    \node[vertex] (Q10) at (-1.5,-3) 	{$V'[2]$}  ;
    \node[vertex] (Q11) at (-0.5,-3) 	{$V'[3]$}  ;
    \node[vertex] (Q12) at (0.5,-3) {$V'[4]$}  ;
    \node[vertex] (Q13) at (1.5,-3)	{$V'[5]$}  ;
    \node[vertex] (Q14) at (2.5,-3) 	{$V'[6]$}  ;
    \node[vertex] (Q15) at (3.5,-3) 	{$V'[7]$}  ;

    \path (Q1) edge node[left]         {$false$}  	(Q2);
    \path (Q1) edge node[right]         {$true$}  	(Q3);
    
    \path (Q2) edge node[left]         {$false$}  	(Q4);
    \path (Q2) edge node[right]         {$true$}  	(Q5);
    
    \path (Q3) edge node[left]         {$false$}  	(Q6);
    \path (Q3) edge node[right]         {$true$}  	(Q7);

    \path (Q4) edge node[left]         {$false$}  	(Q8);
    \path (Q4) edge node[right]         {$true$}  	(Q9);
    
    \path (Q5) edge node[left]         {$false$}  	(Q10);
    \path (Q5) edge node[right]         {$true$}  	(Q11);
    
    \path (Q6) edge node[left]         {$false$}  	(Q12);
    \path (Q6) edge node[right]         {$true$}  	(Q13);
    
    \path (Q7) edge node[left]         {$false$}  	(Q14);
    \path (Q7) edge node[right]         {$true$}  	(Q15);
    
    \end{tikzpicture}
    \caption{Illustration of nodes index coding.\label{codingIndex}}
\end{figure}

%We follow the notations used in previous works \cite{narodytska2018learning, bessiere2009minimising}.
Let $\mathcal{E} = \{e_0, ..., e_{n-1}\}$ be a set of \emph{training examples}, that is, Boolean valuations of a set $\mathcal{F} = \{f_0, f_1, ..., f_{m-1}\}$ of \emph{features}, and let $\mathcal{E}_0, \mathcal{E}_1, ..., \mathcal{E}_{c-1}$ be a partition of $\mathcal{E}$ into classes.
Note that even if we only consider binary features, we can easily handle non-binary features by encoding them in a binary way \cite{bartnikowski2014effects}.
Features that belong to $x$ categories can be represented by $x$ Boolean features where each one represents the affiliation to one category.
If the categories are ordered, then each Boolean feature can represent the affiliation to a smaller or equal category (see Example \ref{codageNum}).
The second encoding provides constraints of type $\leq$ on numerical features for example.
In the following, we denote by $e[f]$ the Boolean valuation of the feature $f \in \mathcal{F}$ in example $e \in \mathcal{E}$.
A \emph{decision tree} is a binary tree where each node is labelled by a single feature and each leaf is labelled by a single class.
A decision tree is said to be \emph{perfect} if all leaves have the same depth and all internal nodes have two children.
Formally, we denote a perfect decision tree by $T = (V, V')$ where $V \in \mathcal{F}^{2^k-1}$ is the set of internal nodes and $V' \in \{0, 1, ..., c-1\}^{2^k}$ is the set of leaves and where $k$ is the depth of the tree.
We denote by $V[i]$ the $i^{\text{th}}$ node in the tree $T$ and by $V[1]$ the root of the tree.
Then we define $V[i \times 2]$ as the left child of $V[i]$ and $V[i \times 2 + 1]$ as the right child.
In a similar way, if $i \geq 2^{k-1}$, we define the leaf $V'[(i-2^{k-1}) \times 2]$ as the left child of $V[i]$ and the leaf $V'[(i-2^{k-1}) \times 2 + 1]$ as the right child. An illustration of this encoding is depicted by Figure~\ref{codingIndex}.

%TODO: définir "features true for o" where o is an observation.

This way of associating a number to every node and leaf may appear complicated, but it will be useful for our Boolean encoding.
We will use the semantics associated with binary coding of node indexes to obtain compact SAT formulas.

If $T$ is a decision tree, and $\mathcal{E}$ is a set of training examples, we say that $T$ is \emph{consistent} with $\mathcal{E}$, denoted $\mathcal{E} \subseteq T$, if each example $e \in \mathcal{E}$ is correctly classified by $T$.
%Thus, in this paper, our objective is to find a decision tree with a minimal number of nodes such that $T$ remains consistent with $\mathcal{E}$.

\begin{example}\label{codageNum}
Let $\mathcal{E}$ be a set of training examples where each example has a single integer feature $f$.
Let $\mathcal{E}_0 = \{ (1), (3) \}$ and $\mathcal{E}_1 = \{ (4), (5) \}$ be the partition of $\mathcal{E}$ into two classes.
Then, we can transform $\mathcal{E}_0$ and $\mathcal{E}_1$ into $\mathcal{E}'_0$ and $\mathcal{E}'_1$ such that each example has four Boolean features $f'_0, f'_1, f'_2, f'_3$.
If the feature $f'_0$ is true, it means that the example is smaller or equal to $1$ for feature $f$;
if the feature $f'_1$ feature is true, it that the example is smaller or equal to $3$ for feature $f$, etc.
Thus, with this transformation we obtain $\mathcal{E}'_0 = \{ (true, false, false, false), (true, true, false, false) \}$ and $\mathcal{E}'_1 = \{ (true, true, true, false), (true, true, true, true) \}$.
\end{example}

\section{Passive Inference}

There are two types of methods for solving the problem of inferring a decision tree from examples.

One group constitutes heuristic methods which try to find a relevant feature for each internal node in polynomial time \cite{breiman1984classification, salzberg1994c4, quinlan1986induction, kass1980exploratory}.
They are often used in practice because of their efficiency; however, they provide no guarantee of optimality.
Thus, better choices in the feature order can lead to smaller decision trees that are consistent with training examples.

Another group includes exact algorithms to determine a decision tree with a minimal number of nodes.
It is a much more complicated problem, as it is NP-complete \cite{laurent1976constructing}.
There are essentially two works that focus on this problem \cite{narodytska2018learning, bessiere2009minimising}.

We propose a SAT formulation that differs from them.
Our approach has two main steps.
In the first step, we seek a perfect decision tree of a minimal depth.
In a second step, we add constraints to reduce the number of nodes in order to potentially obtain an imperfect decision tree.

\subsection{Inferring perfect decision trees of fixed maximum depth}\label{minimize depth}

Our SAT encoding to infer decision trees of a fixed maximum depth is based on the way the nodes are indexed.
As mentioned in Section~\ref{section:def}, the index of a node depends on its position in the tree.
In particular, the root node corresponds to the node $V[1]$, and for each node $V[i]$, the left child corresponds to $V[i \times 2]$ and the right child to $V[i \times 2 + 1]$.
This coding has a useful capability of providing precise information on the position of a node based on the binary coding of its index.
Indeed, reading the binary coding of a node from the highest to the lowest weight bit indicates which branches to take when moving from the root to the node.
For example, if the binary coding of $i$ is 1011, then the node $V[i]$ is reached by taking the right branch of the root, then the left branch, and finally twice the right branch.
Note that if a node $V[i]$ is a descendant of the right branch of a node $V[j]$ then we say that $V[j]$ is a right ancestor of $V[i]$.

The idea of our coding consists in arranging the training examples in the leaves of the tree while respecting the fact that all training examples placed in the same leaf must belong to the same class.
Moreover, if an example is placed in a leaf, then all the right ancestors of that leaf can only be labelled by features true for that example (and conversely for the left ancestor).

The encoding idea is formalized using the following types of Boolean variables.
\begin{itemize}

    \item $X_{i, j}$: If the variable $X_{i, j}$ is true, it means that the example $e_i$ is assigned to a leaf that is a right ancestor of a node located at depth $j$. If $X_{i, j}$ is false, then $e_i$ is assigned to a leaf that is a left ancestor of that node.
    Note that with this semantics on the variables $X_{i, j}$, we have the property that the binary coding $(X_{i,0} X_{i,1} ... X_{i,k-1})$, denoted $X_i$, corresponds to the index of the leaf where the example $e_i$ belongs, i.e., if $X_i = v$, then the example $e_i$ is assigned to the leaf $V'[v]$.
    We also denote by $X_i[..a]$ the number formed by the binary coding of $(X_{i,0} X_{i,1} ... X_{i,a})$.
    
    \item $F_{i, j}$: If $F_{i,j}$ is true, it means that the node $V[i]$ is labelled by the feature $f_j$.
    
    \item $C_{i, j}$: If $C_{i, j}$ is true, it means that the leaf $V'[i]$ is labelled by the class $j$.
    
\end{itemize}

We then use the following set of clauses to formulate constraints that a perfect decision tree of depth $k$ should satisfy.

\noindent
For each $i \in [1, 2^k-1]$, we have the clauses:
\begin{equation}
		\label{equation 1}
	F_{i, 0} \vee F_{i, 1} \vee ... \vee F_{i, m-1}
\end{equation}
These clauses mean that each node should have at least one feature.\\

\noindent
For each $i \in [1, 2^k-1]$ and every features $f_1, f_2$ such that $0 \leq f_1 < f_2 < m$, we have the clauses:
\begin{equation}
		\label{equation 2}
		\neg F_{i, f_1} \vee \neg F_{i, f_2}
\end{equation}
These clauses mean that each node has at most one feature.\\

\noindent
For every $i$ and $f$ such that $e_i[f] = 0$, and each $j \in [0, k-1]$, we have:
\begin{equation}
		\label{equation 3}
        X_{i, j} \Rightarrow \neg F_{X_i[..j], f}
\end{equation}
And for every $i$ and $f$ such that $e_i[f] = 1$, and each $j \in [0, k-1]$, we have:
\begin{equation}
		\label{equation 3 bis}
        X_{i, j} \Rightarrow F_{X_i[..j], f}
\end{equation}
These formulas add constraints that some features cannot be found in certain nodes depending on where the training examples are placed in the decision tree.
We use the binary coding of the index of a leaf to determine which nodes in the tree are its parents.
Thus, all the parent nodes for which the left branch has been taken cannot be labelled by features that must be true, and vice versa.
Note that it is not trivial to translate these formulas into clauses, but we show in Algorithm~\ref{algoGen3} how it could be done.
This algorithm performs a depth-first search of the perfect decision tree in a recursive way.
The variable $q$ corresponds to the index of the current node and $\neg clause$ constraints such that $X_i$ correspond to the index of a $q$ successor.
Each time the algorithm visits a state $q$, it adds constraints on the features that can be labeled by this node based on where $e_i$ is placed in the left or right branch of $q$.
If $e_1$ is placed in the left branch, then $q$ cannot contain a feature that is true for $e_i$ and vice versa with the right branch.\\

\noindent
For each $e_i \in \mathcal{E}_a$ with $a \in [0, c-1]$ and each integer $v \in [0, 2^k-1]$, we have:
\begin{equation}
	\label{equation 4}
	X_i = v \Rightarrow C_{v, a}
\end{equation}
And for each $e_i \in \mathcal{E}_a$, each integer $v \in [0, 2^k-1]$, and $a' \neq a$ we have:
\begin{equation}
	\label{equation 4 bis}
	X_i = v \Rightarrow \neg C_{v, a'}
\end{equation}
These formulas assign the classes to the leaves according to the places of the training examples in the decision tree.
Again, since it is not trivial to translate these formulas into clauses, we show in Algorithm~\ref{algoGen4} how it could be performed.
This algorithm performs a depth-first search of the perfect decision tree such that when it reaches a leaf, $\neg clauses$ corresponds to the index of that leaf.
After that, for each leaf, the algorithm generates the constraints that if $e_i$ is present in that leaf, then that leaf must have the same class as $e_i$.

\begin{algorithm}{\textbf{Input:} A new example $e_i$, a clause $clause$, an index node $q$, the depth of the tree $lvl$ already considered. Initially, $clause=\emptyset$, $q=1$ and $lvl=0$.\\
\textbf{Output:} Clauses for formulas (\ref{equation 3}) and (\ref{equation 3 bis}) when we consider a new example $e_i$}
    \begin{algorithmic}[1]
        \STATE $result = \emptyset$
        
        \IF{$lvl = k$}
            \RETURN $result$
        \ENDIF
        
        \FORALL{$f \in [0, m-1]$ such that $e_i[f] = true$}
            \STATE $result := result \wedge (clause \vee X_{i, lvl} \vee \neg F_{q, f} )$
        \ENDFOR
        
        \STATE $result := result \wedge GenerateFeatureConstraints(e_i, (clause \vee X_{i, lvl}), q \times 2, lvl + 1)$

        \FORALL{$f \in [0, m-1]$ such that $e_i[f] = false$}
            \STATE $result := result \wedge (clause \vee \neg X_{i, lvl} \vee F_{q, f} )$
        \ENDFOR
        
        \STATE $result := result \wedge GenerateFeatureConstraints(e_i, (clause \vee \neg X_{i, lvl}), q \times 2 + 1, lvl + 1)$

        \RETURN $result$
    \end{algorithmic}
    \caption{($GenerateFeatureConstraints$)\label{algoGen3}}%Return an FSM with a most $n$ states consistent with $\mathcal{T}$ if it exists}
\end{algorithm}

\begin{algorithm}{\textbf{Input:} A new example $e_i \in \mathcal{E}_a$, a clause $clause$, a node number $q$, an integer $lvl$ and an integer $lvlMax$.
Initially, $clause = \emptyset$, $q=0$ and $lvl=0$.
\\
\textbf{Output:} Clauses for formulas (\ref{equation 4}) and (\ref{equation 4 bis}) when we consider a new example $e_i$}
    \begin{algorithmic}[1]
                \IF{$lvl = lvlMax$}
                    \STATE $result = (clause \vee C_{q, a})$
                    \FOR{$\mathcal{E}_{a'} \neq \mathcal{E}_{a}$}
                        \STATE $result := result \wedge (clause \vee \neg C_{q, a'})$
                    \ENDFOR
                    \RETURN $result$
                \ENDIF
                
                \RETURN $GenerateClassConstraints(e_i, clause \vee X_{i, lvl}, q \times 2, lvl+1, lvlMax ) \cup GenerateClassConstraints(e_i, clause \vee \neg X_{i, lvl}, q \times 2 + 1, lvl+1, lvlMax )$

        %\EndProcedure
    \end{algorithmic}
    \caption{($GenerateClassConstraints$)\label{algoGen4}}%Return an FSM with a most $n$ nodes consistent with $\mathcal{T}$ if it exists}
\end{algorithm}

\newpage
\subsection{Minimizing the number of nodes}\label{minimize nodes}

Perfect decision trees are often considered as unnecessarily too large, i.e. they can contain too many nodes.
For example, there may be an imperfect decision tree consistent with training examples, with the same depth $k$ as the perfect tree, but with fewer nodes.

In order to find a tree with a minimum number of nodes, we show in this section how we can add constraints to set a maximum number of nodes of the tree.
The idea is to limit the number of leaves that can be assigned to a class in the perfect tree.
Indeed, if a leaf is not assigned to a class, then the parent of this leaf can be replaced by its other child.
By applying this algorithm recursively until all leaves are assigned to a class, we get a decision tree with exactly one leaf more than the internal nodes.
Thus, limiting the number of nodes to $MaxNodes$ has the same effect as limiting the number of leaves to $\lfloor MaxNodes / 2 \rfloor + 1$.

%Thus, for each leaf not assigned to a class, an internal node of the tree as well as this leaf can be deleted. 
%Indeed, a node for which one of his children leads to no labelled nodes is an unnecessary node and can easily be removed from the tree.

To add the constraint of the maximum number of leaves that can be assigned to a class, we add two types of additional variables.
The variables $U_{i}$ which are true if a class is assigned to the leaf $i$, and the variables $H_{i,0}, H_{i,1}, ..., H_{i,MaxNodes+1}$ which will be used to count, with unary coding, the number of leaves labelled by a class.
The variable $H_{i+1, j}$ will be true if there are at least $j$ leaves labelled by a class among the $i$ first leaves.

The clauses encoding the new constraint are as follows:

\noindent
For each $i \in [0, 2^{k}-1]$ and each class $a \in [0, c-1]$, we have the clauses:
\begin{equation}
		\label{equation A}
		\neg C_{i, a} \vee U_{i}
\end{equation}
These clauses assign $U_i$ to true if the leaf $i$ is labelled by a class.\\

\noindent
For each $i \in [0, 2^{k}-1]$ and each class $j \in [0, MaxNodes+1]$, we have the clauses:
\begin{equation}
		\label{equation B}
		\neg H_{i, j} \vee H_{i+1, j}
\end{equation}
These clauses propagate the fact that if $H_{i, j}$ is true, then $H_{i+1, j}$ is also true.

\noindent
For each $i \in [0, 2^{k}-1]$ and each class $j \in [0, MaxNodes+1]$, we have the clauses:
\begin{equation}
		\label{equation C}
		\neg U_{i} \vee \neg H_{i, j} \vee H_{i+1, j+1}
\end{equation}
These clauses increase the value of $H_{i+1}$ by one if $U_{i}$ is true.
Thus $H_{i+1, j}$ is true if there is at least $j$ leaves labelled by class among the $i$ first leaves.\\

\noindent
Finally, we assign the start and end of the counter $H$ :
\begin{equation}
		\label{equation D}
		\neg H_{2^{k+1}, \lfloor MaxNodes / 2 \rfloor + 2} \wedge H_{0,0}
\end{equation}
The first assignment prohibits having more than $ \lfloor MaxNodes / 2 \rfloor + 1 $ leaves, so $MaxNodes$ nodes.
The second assignment sets the counter to $0$.

\begin{proposition}
    The formula for inferring a decision tree of depth $k$ (and a specific number of nodes) from $n$ training examples with $m$ features classed in $c$ classes require $O(2^k \times (n + m + c))$ literals, and $O( 2^k \times (m^2 + m \times n + c) )$ clauses.
\end{proposition}

It can be noted that the number of literals and clauses whether the maximum number of nodes is specified or not is of the same orders of magnitude.
However, finding a tree with a minimum number of nodes will take more time because it requires us to search for this number by a dichotomous search.

\section{Incremental Inference}

To alleviate the complexity associated with large sets of training examples, we propose an approach which, instead of attempting to process all the training examples $\mathcal{E}$ at once, iteratively infers a decision tree from their subset (initially it is an empty set) and uses active inference to refine it when it is not consistent with one of the training examples.
While active inference usually uses an oracle capable of deciding to which class an example belongs, we assign this role to the training examples $\mathcal{E}$.
Even if such an oracle is restricted since it cannot guess the class for all possible input features, nevertheless, as we demonstrate, it leads to an efficient approach for passive inference from training examples.
The approach is formalized in Algorithm \ref{algoLDT}.

\begin{algorithm}{\textbf{Input:} The maximum depth $k$ of the tree to infer, the maximal number $MaxNodes$ of nodes of the tree to infer, the set of training examples $\mathcal{E} = \{\mathcal{E}_0, \mathcal{E}_1, ..., \mathcal{E}_{c-1}\}$.\\
\textbf{Output:} A decision tree consistent with $\mathcal{E}$ with at most $MaxNodes$ nodes and with maximum depth $k$ if it exists.}
    \begin{algorithmic}[1]
                \STATE $C := $ formulas (1) and (2)
                \WHILE{$C$ is satisfiable}
                    \STATE Let $T$ be a decision tree of a solution of $C$
                    \IF{ $\mathcal{E} \subseteq T$ }
                        \RETURN $T$
                    \ENDIF
                    \STATE Let $e \in \mathcal{E}_a$ be an example mislabelled by $T$
                    \STATE $C := C \wedge GenerateFeatureConstraints(e, \emptyset, 1, 0) \wedge GenerateClassConstraints(e, \emptyset, 0, 0, k, a) $
                    \IF{$MaxNodes$ is defined}
                        \STATE $C := C \wedge C'$ where $C'$ is clauses described by formulas (\ref{equation A}), (\ref{equation B}), (\ref{equation C}) and (\ref{equation D}).                 
                    \ENDIF
                \ENDWHILE
                \RETURN ``No solution"
    \end{algorithmic}
    \caption{($InferDecisionTree$)\label{algoLDT}}%Return an FSM with a most $n$ states consistent with $\mathcal{T}$ if it exists}
\end{algorithm}

An illustration of the execution of this algorithm is given in Appendix on a simple example.

\section{Benchmarks}

In this paper, we have presented two algorithms to solve two different problems.

The first algorithm, denoted $DT\_depth$, finds a perfect decision tree of minimal depth.
It uses Algorithm~\ref{algoLDT} without defining $MaxNodes$.
We initially set $k=1$ and while the algorithm is not finding a solution, we increase the value of $k$.

The second algorithm, denoted $DT\_size$, minimizes the depth of the tree and the number of nodes.
It starts by applying $DT\_depth$ to leanr the minimum depth $k$ required to find a decision tree consistent with the training examples.
Then it performs a dichotomy search on the number of nodes allowed between $1$ and $2^{k+1}-1$ to find a decision tree with a minimal number of nodes.

We compare our algorithms with the one of Bessiere et al. \cite{bessiere2009minimising}, denoted $DT2$, and the one of Naradytska et al. \cite{narodytska2018learning}, denoted $DT1$. % \footnote{An implementation is available at \url{http://florent.avellaneda.free.fr/DT.tar.xz}}

The main metric we will compare is the execution time and accuracy.
The accuracy is calculated with a \emph{$k$-fold cross-validation} defined as follows.
We divide the dataset into $k$ equal parts (plus or minus one element).
Then $k-1$ of these parts are used to infer a decision tree, the last part is used to calculate the percentage of its elements correctly classified by the decision tree.
This operation is performed $k$ times to try all possible combinations among these ten parts and the average percentage is calculated.

The prototype was implemented in C++ calling the SAT solver MiniSAT \cite{minisat} and we run the prototype on Ubuntu in a computer with 12GB of RAM and i7-2600K processor.

\subsection{The ``Mouse" dataset}

Our first experiment is performed on the ``Mouse" dataset that the authors Bessiere et al. shared with us.
This dataset has the advantage of having been used with both algorithm $DT1$ and $DT2$.
In Table~\ref{mouse table}, we compare the time and accuracy for different algorithms.
Each entry in rows $DT\_size$ and $DT\_depth$ corresponds to the average over 100 runs.
The first four columns correspond to inferring a decision tree from the whole dataset.
The last column corresponds the 10-fold cross-validations.

\begin{table}[H]\center
  {\caption{Benchmark for ``Mouse" dataset.\label{mouse table}}}
  {\begin{tabular}{|c||c|c|c|c|c|c|}
\hline
Algo & ~Time (s)~ & ~Examples used~ & ~Clauses~ & ~~k~~ & ~Nodes~ & ~acc.~ \\
\hline
\hline
$DT2$ &   577  & 1728 & 3.5M & - & 15 & - \\
$DT1$ &   12.9 & 1728 & 1.2M & - & 15 & - \\
$DT\_{size}$ &  0.075 & 37 & 42K & 4 & 15 & 84.0\% \\
$DT\_{depth}$ & \textbf{0.015} & 33 & 38K  & 4 & 31 & \textbf{85.6\%} \\
\hline
\end{tabular}
}
\end{table}

By analyzing Table~\ref{mouse table}, we can notice that our incremental approach is very effective on this dataset.
Only 37 examples for $DT\_size$ and 33 examples for $DT\_depth$ were used to build an optimal decision tree consistent with the entire dataset.
Thus, thanks to our incremental approach and an efficient SAT formulation, our algorithms are much faster than $DT2$ and $DT1$.
We could not compare the accuracy because this data is missing in the two respective papers for $DT2$ and $DT1$.

\subsection{The ``Car" dataset}

Another data set provided to us and used by the authors Bessiere et al. and Naradytska et al. is ``Car".
This dataset is much more complicated and to the best of our knowledge, no algorithm has been able to infer an optimal decision tree consistent with the entire dataset.
The authors Bessiere et al. process this dataset using linear programming (denoted $DT3$) to minimize the number of nodes.
However, they do not guarantee that the decision tree they find is optimal.
The approach used by the authors Naradytska et al. simplifirs the dataset by considering only 10\% of the data.
Thus, they can infer an optimal decision tree consistent with the 10\% of the data selected in 684 sec.
Table~\ref{car table} compares the results of different algorithms.
Each entry in rows $DT\_size$ and $DT\_depth$ corresponds to the average over ten runs.
The first four columns correspond to inferring a decision tree from the whole dataset.
The last column corresponds the 10-fold cross-validations.

\begin{table}[H]\center
  {\caption{Benchmark for ``Car" dataset.\label{car table}}}
  {\begin{tabular}{|c||c|c|c|c|c|c|}
\hline
Algo & ~Time (s)~ & ~Examples used~ & ~~k~~ & ~Nodes~ & ~acc.~ \\
\hline
\hline
$DT1$ &  684  & 173 &  - & 23.67 & 55\% \\
$DT\_{size}$ & 260 & 758 & 8 & 136 & \textbf{98.8\%} \\
$DT\_{depth}$ & \textbf{170} & 635 & 8 & 511 & \textbf{98.8\%} \\
\hline
$DT3$ & 44.8 & 1729 & - & 92 & 99.22\% \\
\hline
\end{tabular}
}
\end{table}

We can see in Table~\ref{car table} that of the $1729$ examples in the ``Car" dataset, our incremental approach uses less than half of it.
Although this number is still much higher than the number of examples used by the algorithm $DT1$, we can see that our algorithms run faster.
Moreover, since our algorithms ensure that the resulting decision trees are consistent with all training examples, we can see that the accuracy remains very high compared to the $DT1$ algorithm which randomly considers only $10\%$ of training examples.
Note that the algorithm $DT3$ performs better that our algorithm but their algorithm is a heuristic that infers a decision tree without any guarantee of optimality.
In addition, because $DT3$ do not have an optimality constraint, it seek to minimize the number of nodes of a general decision tree without constraint on the depth of the tree.
This way, $DT3$ find trees with fewer nodes, but deeper than ours.

\subsection{Other datasets}

As we mentioned in the introduction of the paper, there is a series of algorithms that addresses a different but very similar problem to ours: inferring optimal decision tree with a given depth such that the total classification error on the training examples is minimized.
In this section, we compare our results against these algorithms.
The datasets we use are extracted from the paper of Verwer and Zhang \cite{verwer2019learning} and are available at \emph{https://github.com/SiccoVerwer/binoct}.
Each dataset corresponds to a 5-fold cross-validation.
In their paper, Verwer and Zhang compare their approach $BinOCT^*$ to two other approaches.
The first one is $CART$ \cite{breiman1984classification}, run from sciki-learn with its default parameter setting but with a fixed maximum depth of the trees generated, and the second one is OCT from Bertsimas and Dunn \cite{bertsimas2017optimal}.
The time limit used is 10~minutes for $BinOCT^*$ and 30 minutes to 2 hours for $OCT$.
The depth of tree used is between $2$ and $4$, but we report in Table~\ref{other table} the best value among the three depths tried.

\begin{table}[H]\center
  {\caption{Benchmark comparing algorithms $DT\_depth$, $DT\_size$, $BinOCT^*$, $CART$ and $OCT$.\label{other table}}}
  {\begin{tabular}{|c||c|c|c|c|c|c|c|c|c|}
\hline
 & \multicolumn{3}{|c|}{$DT\_depth$} & \multicolumn{3}{|c|}{$DT\_size$} & $BinOCT^*$ & $CART$ & $OCT$ \\
Dataset & time (s) & acc. & k & time (s) & acc. & n & acc. & acc. & acc. \\
\hline
\hline
iris & 0.012 & 97.9\% & 3 & 0.019 & 95.8\% & 10.6 & \textbf{98.4\%} & 92.4\% & 93.5\% \\
Monks-probl-1 & 0.016 & 89.0\% & 4.4 & 0.12 & \textbf{92.3}\% & 17 & 87.1\% & 76.8\% & 74.2\% \\
Monks-probl-2 & 0.18 & 68.4\% & 5.8 & 8.6 & \textbf{73.0\%} & 47.8 & 63.3\% & 63.3\% & 54.0\% \\
Monks-probl-3 & 0.02 & 78.7\% & 4.8 & 0.2 & 81.9\% & 23.4 & 93.5\% & \textbf{94.2\%} & \textbf{94.2\%} \\
wine & 0.6 & 88.4\% & 3 & 1.6 & 92.0\% & 7.8 & 92.0\% & 88.9\% & \textbf{94.2\%} \\
balance-scale & 55 & 92.6\% & 8 & 350 & \textbf{92.6\%} & 206 & 78.9\% & 77.5\% & 71.6\% \\

\hline
\hline
Average &  & 85.8\% &  &  & \textbf{87.9\%} &  & 85.5\% &  82.18\% & 81.1\% \\

\hline
\end{tabular}
}
\end{table}

It should be noted that only 6 of the 16 datasets present in the Verwer and Zhang paper could be executed.
The reason is that the decision trees consistent with some datasets are too large and deep to be inferred.
In contrast to the algorithms to which we compare ours, we cannot set a maximum tree depth value because all examples must be correctly classified by the tree we infer.

Note that in Table~\ref{other table}, our algorithms $DT\_depth$ and $DT\_size$ are very fast even when the trees to be inferred are large.
In fact, for the dataset ``balance-scale", our algorithms infer decision trees of depth $8$ in a few minutes while the other algorithms require more time for trees of depth $4$.

Concerning the accuracy of the trees we infer, it seems that when the depth is small ($< 5$) accuracy is equal for all approaches.
However, when the depth becomes bigger, then our algorithms get higher accuracy.
The most obvious example is the dataset balance-scale where we get $92.6\%$ accuracy compared to $78.9\%$ for $BinOCT^*$.

\subsection{Artificial dataset}

\begin{figure}[h]
    \begin{minipage}[c]{.46\linewidth}
        \centering
        \includegraphics[width=1\textwidth]{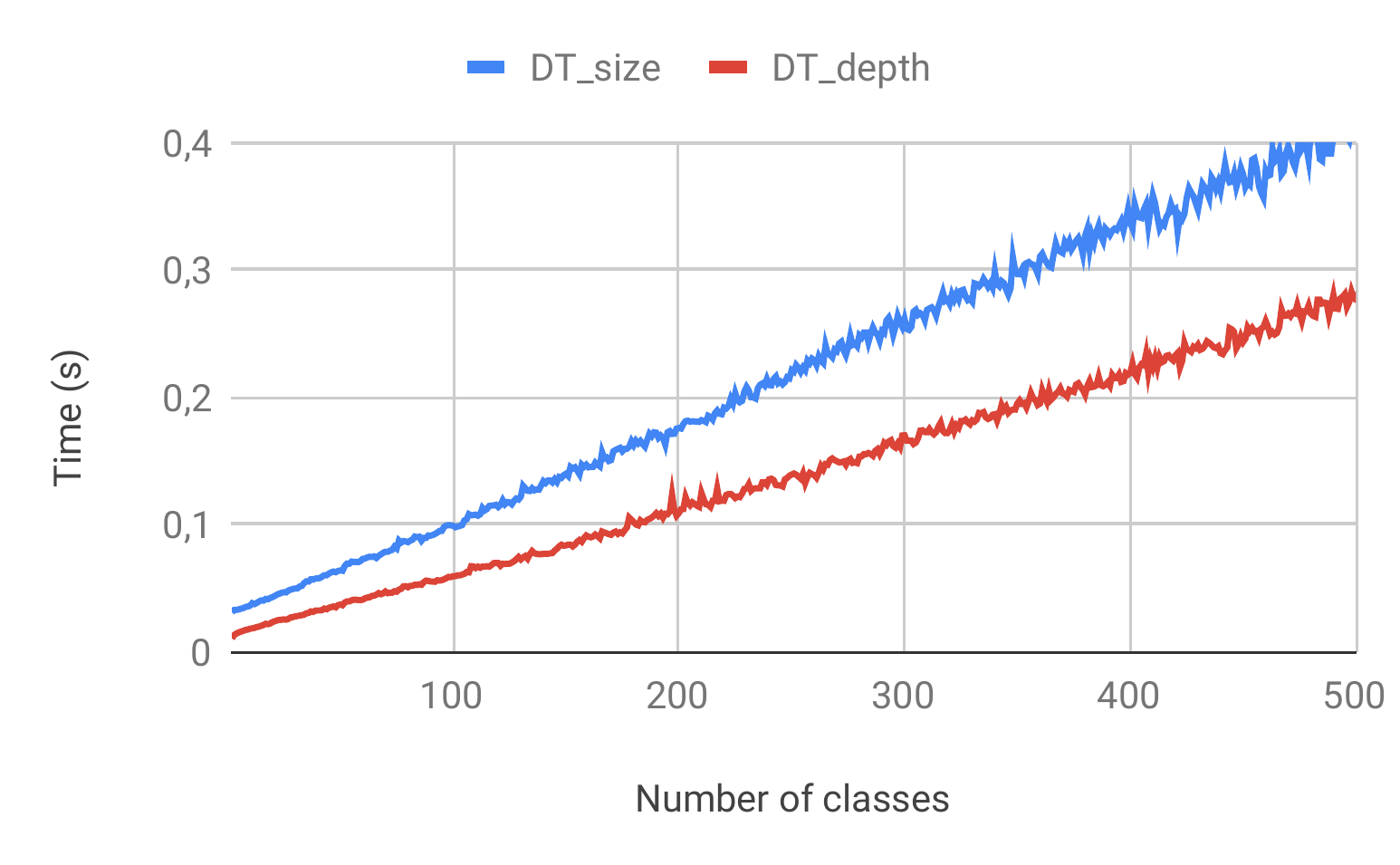}
        \caption{Chart of the average time over 100 runs for $k=4$ and $f=10$.\label{chartC}}
    \label{fig:my_label}
    \end{minipage}
    \hfill%
    \begin{minipage}[c]{.46\linewidth}
        \centering
        \includegraphics[width=1\textwidth]{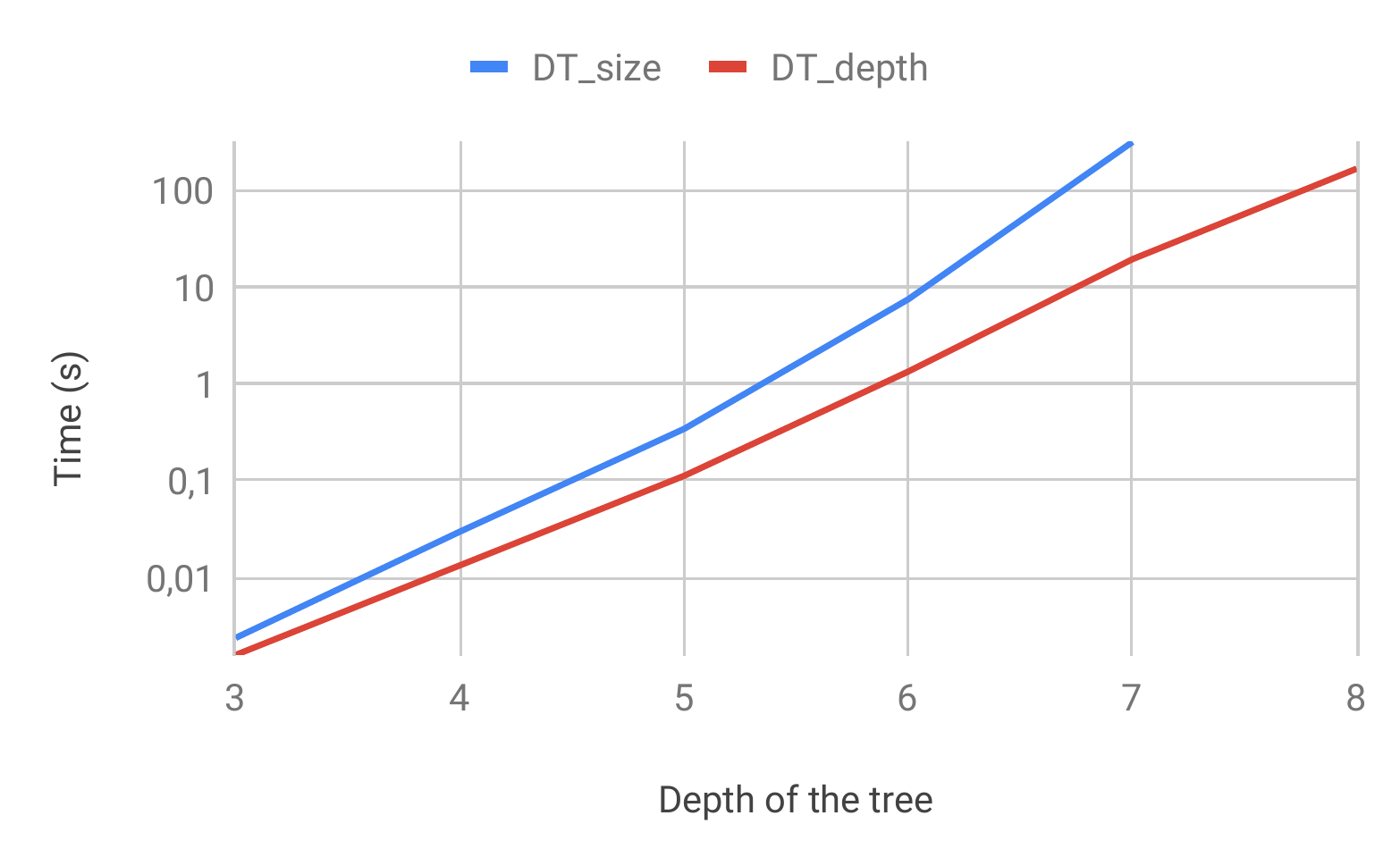}
    \caption{Chart of the average time over 100 runs for $c=2$ and $f=10$.\label{chartK}}
    \end{minipage}

    \begin{minipage}[c]{.46\linewidth}
        \centering
        \includegraphics[width=1\textwidth]{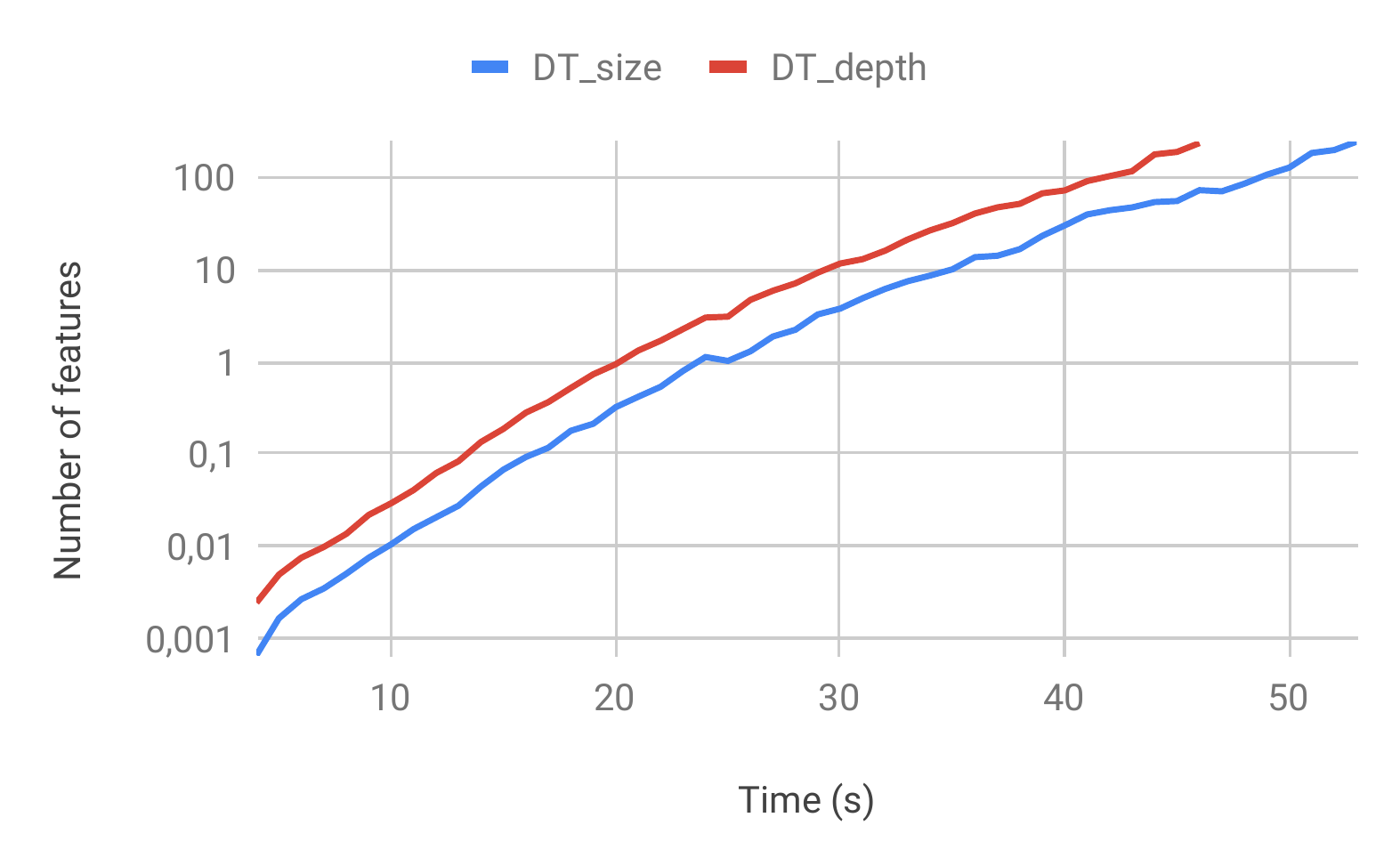}
    \caption{Chart of the average time over 100 runs for $c=2$ and $k=4$.\label{chartF}}
    \end{minipage}
    \hfill%
    \begin{minipage}[c]{.46\linewidth}
        \centering
        \includegraphics[width=1\textwidth]{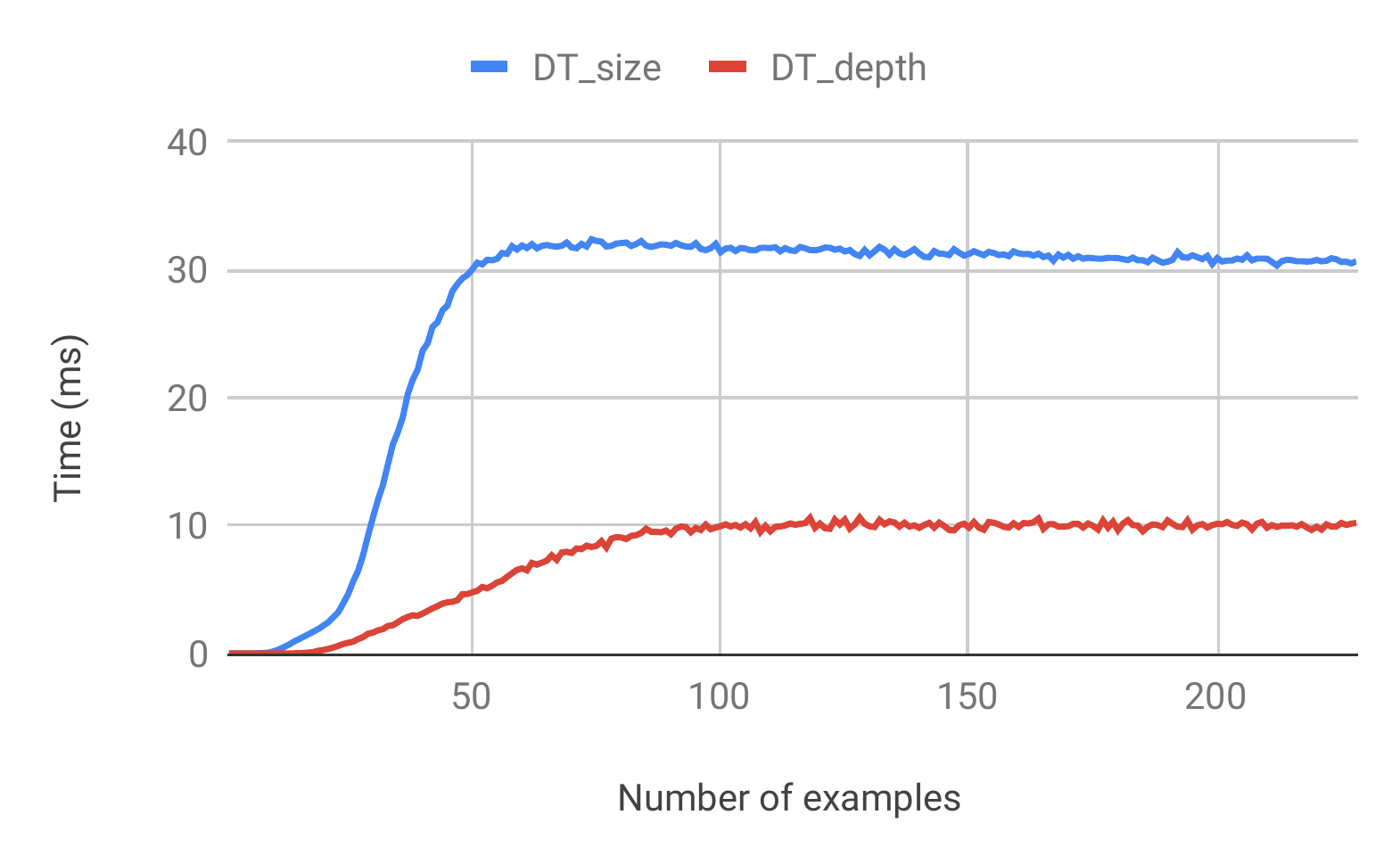}
        \caption{Chart of the average time over 10000 runs for $k=4$, $c=2$ and $f=10$.\label{chartN}}
    \label{fig:my_label}
    \end{minipage}
\end{figure}

A last series of experimentations was carried out in order to see how the execution times of our two algorithms changes with different parameters of the datasets.
The parameters we evaluate are the depth $k$ of the tree to infer, the number $f$ of features, the number $c$ of classes and the number $n$ of training examples.
In order to perform the experiment, we randomly generate decision trees with the characteristics of depths, number of features and number of classes desired, then we randomly generate training examples from such trees.
In Figures~\ref{chartC}, \ref{chartK}, \ref{chartF} and \ref{chartN}, we set three of these parameters and vary the remaining one to observe the effect on execution time.
%Commentaire d'Hilda : On veut bien dire qu'on observe le comportement du temps? 

In Figure~\ref{chartC}, time seems to grow linearly.
The algorithm $DT\_size$ appears to have a coefficient of $1$ which means that the fact that the algorithm is able to infer a decision tree that contains several classes does not bring any gain (or loss) in performance compared to the method consisting of inferring only decision trees with two classes and that would infer $c$ decision trees (one for each class).
However, the algorithm $DT\_depth$ has a coefficient lower than $1$. Thus, the algorithm's ability to infer decision trees with multiple classes provides a performance gain in this case.
In Figure~\ref{chartK}, time grows exponentially with $k$. However, a tree has generally exponentially more nodes than the depth of the tree.
So, the figure indicates that the inference time increases almost polynomially with the number of nodes of the inferred tree.
In Figure~\ref{chartF}, time grows almost exponentially with the number of features.
It is thus the number of features that seems to have the most impact on the inferring time.
Then one way to improve our method is to try to reduce this impact.
In Figure~\ref{chartN}, we observe that time is growing rapidly until it reaches its peak.
This peak corresponds to the number of examples that the algorithm needs to infer a decision tree consisting with all training examples.
Thus, adding more examples will not affect the inference time unlike the previous approaches using SAT solver.

\newpage
\section{Conclusion}

We have presented a method that can infer an optimal decision tree for two definitions of optimality.
The first definition a decision tree of minimal depth and consistent with the training examples is optimal.
The second definition of optimality adds the constraint that the tree, in addition to having a minimum depth, must also have a minimum number of nodes.
Although this optimal decision tree inference problem is known to be NP-complete \cite{laurent1976constructing, hancock1996lower}, we proposed an effective method to solve it.

Our first contribution is an effective SAT formulation that allows us to infer perfect decision trees for a fixed depth consistent with training examples.
%This formulation is based on a binary coding of the positions of the nodes in the tree.
We have also shown how to add constraints in order to set the maximum number of nodes.
In this case, the inferred decision tree will no longer necessarily be a perfect tree.

Our second contribution addresses the scalability issue.
Indeed, the previous approach using SAT solver has the disadvantage that the execution time increases significantly with the number of training examples \cite{narodytska2018learning, bessiere2009minimising}.
Thus, we  proposed  an  approach  which  does  not process all the examples at once, instead it does it incrementally.
The idea of processing a set of traces incrementally is to consider one example at a time, generate a decision tree and verify that it is consistent with the remaining examples.
If it is not, choose an example that is incorrectly classified by the decision tree, i.e., a counterexample, and use it to refine the decision tree.

We evaluated our algorithms using various experiments and compared the execution time and quality of decision trees with other optimal approaches. % in the literature that also seek to infer optimal decision trees for various definitions of optimality.
Experimental results show that our approach performs better than other approaches, with shorter execution times, better prediction accuracy and better scalability.
In addition, our algorithms have been able to process datasets for which, to the best of our knowledge, there are no other inference methods able of producing optimal models consistent with these datasets.

\bibliographystyle{plain}
\bibliography{bibi}

\newpage
\section*{Appendix}

\subsection*{Illustration of the inferring algorithm}

We present here an illustration of the inferring algorithm of a decision tree for depth 2. We use the following dataset $\mathcal{E}_0 = \{(false, false, true, false),$ $(false, false, false, true),$ $(true, false, true, true),$ $(true, true, true, false)\}$ and $\mathcal{E}_1 = \{(false, true, false, true),$ $(false, true, true, false),$ $(true, false, false, false),$ $(true, true, false, true)\}$

\noindent
\textbf{Initialization:}

\begin{tabular}{ll}
     Formula (1):~&  $(\neg F_{1, 0} \vee \neg F_{1, 1}) \wedge 
                    (\neg F_{1, 0} \vee \neg F_{1, 2}) \wedge 
                    (\neg F_{1, 0} \vee \neg F_{1, 3}) \wedge $\\
                    &$(\neg F_{1, 1} \vee \neg F_{1, 2}) \wedge
                    (\neg F_{1, 1} \vee \neg F_{1, 3}) \wedge 
                    (\neg F_{1, 2} \vee \neg F_{1, 3}) \wedge $\\
                    &$(\neg F_{2, 0} \vee \neg F_{2, 1}) \wedge 
                    (\neg F_{2, 0} \vee \neg F_{2, 2}) \wedge 
                    (\neg F_{2, 0} \vee \neg F_{2, 3}) \wedge $\\
                    &$(\neg F_{2, 1} \vee \neg F_{2, 2}) \wedge 
                    (\neg F_{2, 1} \vee \neg F_{2, 3}) \wedge 
                    (\neg F_{2, 2} \vee \neg F_{2, 3}) \wedge $\\
                    &$(\neg F_{3, 0} \vee \neg F_{3, 1}) \wedge 
                    (\neg F_{3, 0} \vee \neg F_{3, 2}) \wedge 
                    (\neg F_{3, 0} \vee \neg F_{3, 3}) \wedge $\\
                    &$(\neg F_{3, 1} \vee \neg F_{3, 2}) \wedge 
                    (\neg F_{3, 1} \vee \neg F_{3, 3}) \wedge 
                    (\neg F_{3, 2} \vee \neg F_{3, 3})$\\
\end{tabular}

\begin{tabular}{ll}
Formula (2):~& $(F_{1, 0} \vee F_{1, 1} \vee F_{1, 2} \vee F_{1, 3}) \wedge $\\
&$(F_{2, 0} \vee F_{2, 1} \vee F_{2, 2} \vee F_{2, 3} )\wedge $\\
&$(F_{3, 0} \vee F_{3, 1} \vee F_{3, 2} \vee F_{3, 3})$\\
\end{tabular}

\begin{figure}[H]
\center
    \begin{tikzpicture}[scale=1.4,shorten >=1pt,->]
    \tikzstyle{vertex}=[circle,fill=black!10,draw=black!75,minimum size=25pt,inner sep=0pt]

    \node[vertex] (Q1) at (0,0) 	{$f_1?$}  ;
    \node[vertex] (Q2) at (-1,-1) 	{$f_0?$}  ;
    \node[vertex] (Q3) at (1,-1) 	{$f_0?$}  ;
    
    \node[vertex] (Q4) at (-1.5,-2) {$1$}  ;
    \node[vertex] (Q5) at (-0.5,-2)	{$1$}  ;
    \node[vertex] (Q6) at (0.5,-2) 	{$1$}  ;
    \node[vertex] (Q7) at (1.5,-2) 	{$1$}  ;

    \path (Q1) edge node[left]         {$false$}  	(Q2);
    \path (Q1) edge node[right]         {$true$}  	(Q3);
    
    \path (Q2) edge node[left]         {$false$}  	(Q4);
    \path (Q2) edge node[right]         {$true$}  	(Q5);
    
    \path (Q3) edge node[left]         {$false$}  	(Q6);
    \path (Q3) edge node[right]         {$true$}  	(Q7);
    
    \end{tikzpicture}
    \caption{Conjecture}
\end{figure}

\noindent
\textbf{Add example: $e_0 = (0, 0, 1, 0) \in \mathcal{E}_0$}

\begin{tabular}{ll}
Formula (3):~& $(X_{0, 0} \vee \neg F_{1, 2}) \wedge (X_{0, 0} \vee X_{0, 1} \vee \neg F_{2, 2}) \wedge (\neg X_{0, 0} \vee X_{0, 1} \vee \neg F_{3, 2})$\\
\end{tabular}

\begin{tabular}{ll}
Formula (4):~& $(X_{0, 0} \vee \neg X_{0, 1} \vee \neg F_{2, 0}) \wedge (X_{0, 0} \vee \neg X_{0, 1} \vee \neg F_{2, 1}) \wedge$\\
&$(X_{0, 0} \vee \neg X_{0, 1} \vee \neg F_{2, 3}) \wedge (\neg X_{0, 0} \vee \neg F_{1, 0}) \wedge (\neg X_{0, 0} \vee \neg F_{1, 1}) \wedge$\\
&$(\neg X_{0, 0} \vee \neg F_{1, 3}) \wedge (\neg X_{0, 0} \vee \neg X_{0, 1} \vee \neg F_{3, 0}) \wedge$\\
&$(\neg X_{0, 0} \vee \neg X_{0, 1} \vee \neg F_{3, 1}) \wedge (\neg X_{0, 0} \vee \neg X_{0, 1} \vee \neg F_{3, 3})$\\
\end{tabular}

\begin{tabular}{ll}
Formula (5):~& $(X_{0, 0} \vee X_{0, 1} \vee C_{0, 0}) \wedge (X_{0, 0} \vee \neg X_{0, 1} \vee C_{1, 0}) \wedge$\\
&$(\neg X_{0, 0} \vee X_{0, 1} \vee C_{2, 0}) \wedge (\neg X_{0, 0} \vee \neg X_{0, 1} \vee C_{3, 0})$\\
\end{tabular}

\begin{tabular}{ll}
Formula (6):~&$(X_{0, 0} \vee X_{0, 1} \vee \neg C_{0, 1}) \wedge (X_{0, 0} \vee \neg X_{0, 1} \vee \neg C_{1, 1})$\\
&$(\neg X_{0, 0} \vee X_{0, 1} \vee \neg C_{2, 1}) \wedge (\neg X_{0, 0} \vee \neg X_{0, 1} \vee \neg C_{3, 1})$\\
\end{tabular}

\begin{figure}[H]
\center
    \begin{tikzpicture}[scale=1.4,shorten >=1pt,->]
    \tikzstyle{vertex}=[circle,fill=black!10,draw=black!75,minimum size=25pt,inner sep=0pt]

    \node[vertex] (Q1) at (0,0) 	{$f_1?$}  ;
    \node[vertex] (Q2) at (-1,-1) 	{$f_0?$}  ;
    \node[vertex] (Q3) at (1,-1) 	{$f_0?$}  ;
    
    \node[vertex] (Q4) at (-1.5,-2) {$0$}  ;
    \node[vertex] (Q5) at (-0.5,-2)	{$-$}  ;
    \node[vertex] (Q6) at (0.5,-2) 	{$-$}  ;
    \node[vertex] (Q7) at (1.5,-2) 	{$-$}  ;

    \path (Q1) edge node[left]         {$false$}  	(Q2);
    \path (Q1) edge node[right]         {$true$}  	(Q3);
    
    \path (Q2) edge node[left]         {$false$}  	(Q4);
    \path (Q2) edge node[right]         {$true$}  	(Q5);
    
    \path (Q3) edge node[left]         {$false$}  	(Q6);
    \path (Q3) edge node[right]         {$true$}  	(Q7);
    
    \end{tikzpicture}
    \caption{Conjecture}
\end{figure}

\noindent
\textbf{Add example: $e_1 = (1, 0, 1, 1) \in \mathcal{E}_0$}

\begin{tabular}{ll}
Formula (3):~& $(X_{1, 0} \vee \neg F_{1, 0}) \wedge (X_{1, 0} \vee \neg F_{1, 2}) \wedge (X_{1, 0} \vee \neg F_{1, 3}) \wedge$\\
&$(X_{1, 0} \vee X_{1, 1} \vee \neg F_{2, 0}) \wedge (X_{1, 0} \vee X_{1, 1} \vee \neg F_{2, 2}) \wedge $\\
&$(X_{1, 0} \vee X_{1, 1} \vee \neg F_{2, 3}) \wedge (\neg X_{1, 0} \vee X_{1, 1} \vee \neg F_{3, 0}) \wedge$\\
&$ (\neg X_{1, 0} \vee X_{1, 1} \vee \neg F_{3, 2}) \wedge (\neg X_{1, 0} \vee X_{1, 1} \vee \neg F_{3, 3})$\\
\end{tabular}

\begin{tabular}{ll}
Formula (4):~& $(X_{1, 0} \vee \neg X_{1, 1} \vee \neg F_{2, 1}) \wedge (\neg X_{1, 0} \vee \neg F_{1, 1}) \wedge$\\
&$(\neg X_{1, 0} \vee \neg X_{1, 1} \vee \neg F_{3, 1})$\\
\end{tabular}

\begin{tabular}{ll}
Formula (5):~& $(X_{1, 0} \vee X_{1, 1} \vee C_{0, 0}) \wedge (\neg X_{1, 0} \vee X_{1, 1} \vee C_{2, 0}) \wedge$\\
&$(X_{1, 0} \vee \neg X_{1, 1} \vee C_{1, 0}) \wedge (\neg X_{1, 0} \vee \neg X_{1, 1} \vee C_{3, 0})$\\
\end{tabular}

\begin{tabular}{ll}
Formula (6):~& $(X_{1, 0} \vee X_{1, 1} \vee \neg C_{0, 1}) \wedge (X_{1, 0} \vee \neg X_{1, 1} \vee \neg C_{1, 1}) \wedge$\\
&$ (\neg X_{1, 0} \vee X_{1, 1} \vee \neg C_{2, 1}) \wedge (\neg X_{1, 0} \vee \neg X_{1, 1} \vee \neg C_{3, 1}) $\\
\end{tabular}

\begin{figure}[H]
\center
    \begin{tikzpicture}[scale=1.4,shorten >=1pt,->]
    \tikzstyle{vertex}=[circle,fill=black!10,draw=black!75,minimum size=25pt,inner sep=0pt]

    \node[vertex] (Q1) at (0,0) 	{$f_1?$}  ;
    \node[vertex] (Q2) at (-1,-1) 	{$f_0?$}  ;
    \node[vertex] (Q3) at (1,-1) 	{$f_0?$}  ;
    
    \node[vertex] (Q4) at (-1.5,-2) {$0$}  ;
    \node[vertex] (Q5) at (-0.5,-2)	{$0$}  ;
    \node[vertex] (Q6) at (0.5,-2) 	{$-$}  ;
    \node[vertex] (Q7) at (1.5,-2) 	{$-$}  ;

    \path (Q1) edge node[left]         {$false$}  	(Q2);
    \path (Q1) edge node[right]         {$true$}  	(Q3);
    
    \path (Q2) edge node[left]         {$false$}  	(Q4);
    \path (Q2) edge node[right]         {$true$}  	(Q5);
    
    \path (Q3) edge node[left]         {$false$}  	(Q6);
    \path (Q3) edge node[right]         {$true$}  	(Q7);
    
    \end{tikzpicture}
    \caption{Conjecture}
\end{figure}

\noindent
\textbf{Add example: $e_2 = (1, 1, 1, 0) \in \mathcal{E}_0$}

\begin{tabular}{ll}
Formula (3):~& $ (X_{2, 0} \vee \neg F_{1, 0}) \wedge (X_{2, 0} \vee \neg F_{1, 1}) \wedge (X_{2, 0} \vee \neg F_{1, 2}) \wedge $\\
&$ (X_{2, 0} \vee X_{2, 1} \vee \neg F_{2, 0}) \wedge (X_{2, 0} \vee X_{2, 1} \vee \neg F_{2, 1}) \wedge $\\
&$ (X_{2, 0} \vee X_{2, 1} \vee \neg F_{2, 2}) $\\
\end{tabular}

\begin{tabular}{ll}
Formula (4):~& $(X_{2, 0} \vee \neg X_{2, 1} \vee \neg F_{2, 3}) \wedge (\neg X_{2, 0} \vee \neg F_{1, 3}) \wedge$\\
&$(\neg X_{2, 0} \vee \neg X_{2, 1} \vee \neg F_{3, 3})$\\
\end{tabular}

\begin{tabular}{ll}
Formula (5):~& $ (X_{2, 0} \vee X_{2, 1} \vee C_{0, 0}) \wedge (X_{2, 0} \vee \neg X_{2, 1} \vee C_{1, 0}) \wedge $\\
&$ (\neg X_{2, 0} \vee X_{2, 1} \vee C_{2, 0}) \wedge (\neg X_{2, 0} \vee \neg X_{2, 1} \vee C_{3, 0}) $\\
\end{tabular}

\begin{tabular}{ll}
Formula (6):~& $ (X_{2, 0} \vee X_{2, 1} \vee \neg C_{0, 1}) \wedge (X_{2, 0} \vee \neg X_{2, 1} \vee \neg C_{1, 1}) \wedge $\\
&$ (\neg X_{2, 0} \vee X_{2, 1} \vee \neg C_{2, 1}) \wedge (\neg X_{2, 0} \vee \neg X_{2, 1} \vee \neg C_{3, 1}) $\\
\end{tabular}

\begin{figure}[h!]
\center
    \begin{tikzpicture}[scale=1.4,shorten >=1pt,->]
    \tikzstyle{vertex}=[circle,fill=black!10,draw=black!75,minimum size=25pt,inner sep=0pt]

    \node[vertex] (Q1) at (0,0) 	{$f_1?$}  ;
    \node[vertex] (Q2) at (-1,-1) 	{$f_0?$}  ;
    \node[vertex] (Q3) at (1,-1) 	{$f_0?$}  ;
    
    \node[vertex] (Q4) at (-1.5,-2) {$0$}  ;
    \node[vertex] (Q5) at (-0.5,-2)	{$0$}  ;
    \node[vertex] (Q6) at (0.5,-2) 	{$-$}  ;
    \node[vertex] (Q7) at (1.5,-2) 	{$0$}  ;

    \path (Q1) edge node[left]         {$false$}  	(Q2);
    \path (Q1) edge node[right]         {$true$}  	(Q3);
    
    \path (Q2) edge node[left]         {$false$}  	(Q4);
    \path (Q2) edge node[right]         {$true$}  	(Q5);
    
    \path (Q3) edge node[left]         {$false$}  	(Q6);
    \path (Q3) edge node[right]         {$true$}  	(Q7);
    
    \end{tikzpicture}
    \caption{Conjecture}
\end{figure}

\noindent
\textbf{Add example: $e_3 = (0, 1, 0, 1) \in \mathcal{E}_1$}

\begin{tabular}{ll}
Formula (3):~& $ (X_{3, 0} \vee \neg F_{1, 1}) \wedge (X_{3, 0} \vee \neg F_{1, 3}) \wedge$\\
&$ (X_{3, 0} \vee X_{3, 1} \vee \neg F_{2, 1}) \wedge (X_{3, 0} \vee X_{3, 1} \vee \neg F_{2, 3}) \wedge $\\
&$ (\neg X_{3, 0} \vee X_{3, 1} \vee \neg F_{3, 1}) \wedge (\neg X_{3, 0} \vee X_{3, 1} \vee \neg F_{3, 3}) $\\
\end{tabular}

\begin{tabular}{ll}
Formula (4):~& $ (X_{3, 0} \vee \neg X_{3, 1} \vee \neg F_{2, 0}) \wedge (X_{3, 0} \vee \neg X_{3, 1} \vee \neg F_{2, 2} )\wedge $\\
&$ (\neg X_{3, 0} \vee \neg F_{1, 0}) \wedge (\neg X_{3, 0} \vee \neg F_{1, 2}) \wedge $\\
&$ (\neg X_{3, 0} \vee \neg X_{3, 1} \vee \neg F_{3, 0}) \wedge (\neg X_{3, 0} \vee \neg X_{3, 1} \vee \neg F_{3, 2}) $\\
\end{tabular}

\begin{tabular}{ll}
Formula (5):~& $ (X_{3, 0} \vee X_{3, 1} \vee C_{0, 1}) \wedge (X_{3, 0} \vee \neg X_{3, 1} \vee C_{1, 1}) \wedge $\\
&$ (\neg X_{3, 0} \vee X_{3, 1} \vee C_{2, 1)} \wedge (\neg X_{3, 0} \vee \neg X_{3, 1} \vee C_{3, 1}) $\\
\end{tabular}

\begin{tabular}{ll}
Formula (6):~& $ (X_{3, 0} \vee X_{3, 1} \vee \neg C_{0, 0}) \wedge (X_{3, 0} \vee \neg X_{3, 1} \vee \neg C_{1, 0}) \wedge $\\
&$ (\neg X_{3, 0} \vee X_{3, 1} \vee \neg C_{2, 0}) \wedge (\neg X_{3, 0} \vee \neg X_{3, 1} \vee \neg C_{3, 0}) $\\
\end{tabular}

\begin{figure}[H]
\center
    \begin{tikzpicture}[scale=1.4,shorten >=1pt,->]
    \tikzstyle{vertex}=[circle,fill=black!10,draw=black!75,minimum size=25pt,inner sep=0pt]

    \node[vertex] (Q1) at (0,0) 	{$f_1?$}  ;
    \node[vertex] (Q2) at (-1,-1) 	{$f_0?$}  ;
    \node[vertex] (Q3) at (1,-1) 	{$f_0?$}  ;
    
    \node[vertex] (Q4) at (-1.5,-2) {$0$}  ;
    \node[vertex] (Q5) at (-0.5,-2)	{$0$}  ;
    \node[vertex] (Q6) at (0.5,-2) 	{$1$}  ;
    \node[vertex] (Q7) at (1.5,-2) 	{$0$}  ;

    \path (Q1) edge node[left]         {$false$}  	(Q2);
    \path (Q1) edge node[right]         {$true$}  	(Q3);
    
    \path (Q2) edge node[left]         {$false$}  	(Q4);
    \path (Q2) edge node[right]         {$true$}  	(Q5);
    
    \path (Q3) edge node[left]         {$false$}  	(Q6);
    \path (Q3) edge node[right]         {$true$}  	(Q7);
    
    \end{tikzpicture}
    \caption{Conjecture}
\end{figure}

\noindent
\textbf{Add example: $e_4 = (1, 0, 0, 0) \in \mathcal{E}_1$}

\begin{tabular}{ll}
Formula (3):~& $ (X_{4, 0} \vee \neg F_{1, 0}) \wedge (X_{4, 0} \vee X_{4, 1} \vee \neg F_{2, 0}) \wedge (\neg X_{4, 0} \vee X_{4, 1} \vee \neg F_{3, 0}) $\\
\end{tabular}

\begin{tabular}{ll}
Formula (4):~& $ (X_{4, 0} \vee \neg X_{4, 1} \vee \neg F_{2, 1}) \wedge (X_{4, 0} \vee \neg X_{4, 1} \vee \neg F_{2, 2}) \wedge $\\
&$ (X_{4, 0} \vee \neg X_{4, 1} \vee \neg F_{2, 3}) \wedge (\neg X_{4, 0} \vee \neg F_{1, 1}) \wedge (\neg X_{4, 0} \vee \neg F_{1, 2}) \wedge $\\
&$ (\neg X_{4, 0} \vee \neg F_{1, 3}) \wedge (\neg X_{4, 0} \vee \neg X_{4, 1} \vee \neg F_{3, 1}) \wedge $\\
&$ (\neg X_{4, 0} \vee \neg X_{4, 1} \vee \neg F_{3, 2}) \wedge (\neg X_{4, 0} \vee \neg X_{4, 1} \vee \neg F_{3, 3}) $\\
\end{tabular}

\begin{tabular}{ll}
Formula (5):~& $ (X_{4, 0} \vee X_{4, 1} \vee C_{0, 1}) \wedge (X_{4, 0} \vee \neg X_{4, 1} \vee C_{1, 1}) \wedge $\\
&$ (\neg X_{4, 0} \vee X_{4, 1} \vee C_{2, 1}) \wedge (\neg X_{4, 0} \vee \neg X_{4, 1} \vee C_{3, 1}) $\\
\end{tabular}

\begin{tabular}{ll}
Formula (6):~& $ (X_{4, 0} \vee X_{4, 1} \vee \neg C_{0, 0}) \wedge (X_{4, 0} \vee \neg X_{4, 1} \vee \neg C_{1, 0}) \wedge $\\
&$ (\neg X_{4, 0} \vee X_{4, 1} \vee \neg C_{2, 0}) \wedge (\neg X_{4, 0} \vee \neg X_{4, 1} \vee \neg C_{3, 0}) $\\
\end{tabular}

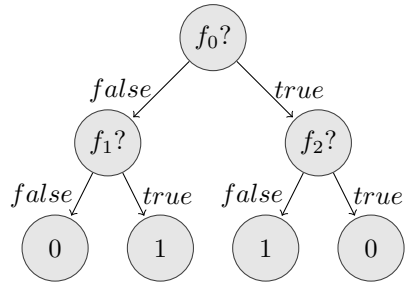
\begin{figure}[H]
\center
    \begin{tikzpicture}[scale=1.4,shorten >=1pt,->]
    \tikzstyle{vertex}=[circle,fill=black!10,draw=black!75,minimum size=25pt,inner sep=0pt]

    \node[vertex] (Q1) at (0,0) 	{$f_0?$}  ;
    \node[vertex] (Q2) at (-1,-1) 	{$f_1?$}  ;
    \node[vertex] (Q3) at (1,-1) 	{$f_2?$}  ;
    
    \node[vertex] (Q4) at (-1.5,-2) {$0$}  ;
    \node[vertex] (Q5) at (-0.5,-2)	{$1$}  ;
    \node[vertex] (Q6) at (0.5,-2) 	{$1$}  ;
    \node[vertex] (Q7) at (1.5,-2) 	{$0$}  ;

    \path (Q1) edge node[left]         {$false$}  	(Q2);
    \path (Q1) edge node[right]         {$true$}  	(Q3);
    
    \path (Q2) edge node[left]         {$false$}  	(Q4);
    \path (Q2) edge node[right]         {$true$}  	(Q5);
    
    \path (Q3) edge node[left]         {$false$}  	(Q6);
    \path (Q3) edge node[right]         {$true$}  	(Q7);
    
    \end{tikzpicture}
    \caption{Final Solution}
\end{figure}

\end{document}